\documentclass[twocolumn]{article}

\usepackage{framed,multirow}
\usepackage{times}

\usepackage{amssymb}
\usepackage{latexsym}

\usepackage{url}
\usepackage{xcolor}
\definecolor{newcolor}{rgb}{.8,.349,.1}

\usepackage{xspace}
\usepackage{amsmath,amsfonts,amssymb}
\usepackage{algpseudocode}
\usepackage{tcolorbox}
\usepackage{verbatim} 
\usepackage{nicefrac} 

\usepackage{subfig}
\usepackage{makecell}
\usepackage{latexsym}
\usepackage{multirow}
\usepackage{adjustbox}
\usepackage{enumitem}
\usepackage{makecell}
\usepackage{algorithm}

\newcommand{\gbar}{\overline{\mathbf{g}}}
\newcommand{\gtilde}{\tilde{\mathbf{g}}}

\newcommand{\VM}{\mathcal{V}}

\newcommand{\Alg}[1]{Algorithm~\ref{#1}}
\newcommand{\Eqn}[1]{Eqn (\ref{#1})}

\newcommand{\Def}[1]{Def \ref{#1}}
\newcommand{\Thm}[1]	{Thm \ref{#1}}

\newcommand{\DirDP}{\textsc{DirDP-SGD}\xspace}
\newcommand{\dpriv}{\textrm{metric DP}\xspace}
\newcommand{\IGA}{IGA\xspace}
\definecolor{greenmunsell}{rgb}{0.0, 0.66, 0.47}

\newtheorem{definition}{Definition}
\newtheorem{theorem}{Theorem}
\newtheorem{corollary}{Corollary}
\newenvironment{proof}{\paragraph{Proof:}}{\hfill$\square$}

\usepackage[utf8]{inputenc}
\usepackage[T1]{fontenc}
\usepackage[english]{babel}
\usepackage{lipsum} 
\usepackage{multicol} 
\usepackage[numbers]{natbib} 

\title{Applying Directional Noise to Deep Learning}
\author{
  Pedro Faustini \and
  Natasha Fernandes \and
  Shakila Tonni \and
  Annabelle McIver \and
  Mark Dras
  \vspace{6pt} \\
  \small Macquarie University \\
  \small \texttt{pedro.arrudafaustini@hdr.mq.edu.au}\\
  \small \texttt{\{natasha.fernandes, shakila.tonni, annabelle.mciver, mark.dras\}@mq.edu.au}
}

\date{}
\begin{document}

\maketitle

\begin{abstract}
Differentially Private Stochastic Gradient Descent (DP-SGD) is a key method for applying privacy in the training of deep learning models.  It applies isotropic Gaussian noise to gradients during training, which can perturb these gradients in any direction, damaging utility. Metric DP, however, can provide alternative mechanisms based on arbitrary metrics that might be more suitable for preserving utility. In this paper, we apply  \textit{directional privacy}, via a mechanism based on the von Mises-Fisher (VMF) distribution, to perturb gradients in terms of \textit{angular distance} so that gradient direction is broadly preserved. We show that this provides both $\epsilon$-DP and $\epsilon d$-privacy for deep learning training, rather than the $(\epsilon, \delta)$-privacy of the Gaussian mechanism. Experiments on key datasets then indicate that the VMF mechanism can outperform the Gaussian in the utility-privacy trade-off. In particular, our experiments provide a direct empirical comparison of privacy between the two approaches in terms of their ability to defend against reconstruction and membership inference.
\end{abstract}

\section{Introduction}
\label{sec:intro}

A popular method for protecting deep learning models against attacks is to train them with Differential Privacy (DP) \citep{dwork-roth:2014}. \citet{song-etal:2013} first proposed adding $\epsilon$-DP noise to the gradient vectors within each batch, thereby providing an $\epsilon$ guarantee over training datasets. They noted, though, that the noise impacted SGD utility significantly. Later developments improved its performance. For example, \citet{abadi-etal:2016:CCS} proposed an accountant with tighter bounds for the privacy budget for the Gaussian mechanism. 

Gaussian noise is isotropic: its noise vector is equally likely to point in any direction in the high-dimensional space of the gradients. In contrast, we might expect that the utility of the model would be better served by a mechanism which is designed to preserve the \emph{direction} of the gradients. Intuitively, the more their directions are preserved, the better is the gradient descent algorithm going to to minimise the loss function.

The idea of tuning the \emph{shape} of the noise arises in the context of \emph{metric differential privacy} or $d$-privacy~\citep{chatzikokolakis-etal:2013:PETS}, a generalisation of DP in which the notion of adjacency is generalised to a distinguishability metric $d$. Metric differential privacy is a unifying definition which subsumes both central and local DP, the former recoverable by choosing $d$ to be the Hamming metric on databases, and the latter by choosing $d$ to be the Discrete metric on individual data points. Importantly, by careful choice of the metric $d$, $d$-privacy mechanisms can provide a better privacy-utility trade-off than with standard DP.


Our \textbf{key idea} is to implement DP-SGD with directional noise, so that with high likelihood a reported gradient is close in direction to the original gradient and further away (in direction) with diminishing likelihood. The aim of the present paper is to show that these kinds of directional privacy mechanisms applied to deep learning training can have less impact on model performance because the application of noise can be more targeted while providing $\epsilon$-privacy guarantees via metric DP.

We compare the Gaussian-noise mechanism of \citet{abadi-etal:2016:CCS} with the (directional privacy) von Mises-Fisher mechanism of  \citet{weggenmann-kerschbaum:2021:CCS}, who applied it to recurrent temporal data rather than deep learning.
A key question which arises when using different types of DP mechanisms is \emph{how to compare the epsilons?} since their guarantees are often not comparable, as noted in, for example, \citet{DBLP:conf/uss/Jayaraman019} and \citet{10.1007/978-3-030-81242-3_2}.  We thus compare them empirically w.r.t. their efficacy at preventing an enhanced MIA of \citet{hu-etal:2022} and a method that reconstructs training data from gradients \citep{geiping-etal:2020:NeurIPS}. We then show that our \textbf{ \DirDP } performs notably better on  major datasets for comparable levels of defence against the aforementioned attacks.

Our contributions are as follows:

\begin{itemize}[noitemsep,topsep=0pt,parsep=0pt,partopsep=0pt,leftmargin=*]
    \item We apply for the first time a metric DP mechanism based on angular distance --- via the von Mises-Fisher (VMF) distribution --- to use as an alternative to standard DP-SGD; 
    
    \item We show that this provides $\epsilon d_\theta$-privacy (for angular distance $d_\theta$) as well as $\epsilon$-DP  for the training as a whole;

    \item We analyse membership inference and gradient-based reconstruction attacks for empirically comparing privacy, and show why using both together is appropriate in this context; and

    \item Given this, we show that overall on three standard datasets, our VMF mechanism outperforms Gaussian noise when defending against attacks.
\end{itemize}

\vspace{-0.1cm}

\section{Related Work}
\label{sec:lit-rev}

\vspace{-0.05cm}

\paragraph{Metric Differential Privacy}
\label{sec:lit-rev-dpriv}

We adopt a relaxation of DP called metric differential privacy, introduced by \citet{chatzikokolakis-etal:2013:PETS} and also known as generalised DP, $d$-privacy,  and $d_{\mathcal{X}}$-privacy. It was first applied to the problem of geo-location privacy \citep{andres2013geo} in which the user reveals an approximate location to receive location-based services. 

In the area of NLP, \citet{fernandes-etal:2019:POST} proposed a \dpriv mechanism for authorship privacy using the Earth Mover's distance as the metric.  
A related application that takes a similar spatial perspective has been to k-means clustering \citep{yang-etal:2022}. None of these is concerned with differentially private training of a deep learner in the manner of DP-SGD.

The application of \dpriv that we draw on is not related to the existing uses in deep learning just described.  In the context of providing privacy guarantees to sleep study data, \citet{weggenmann-kerschbaum:2021:CCS} applied \dpriv to periodic data 
noting that periodicity can be represented as a direction on a circle, and `directional noise' perturbs this direction while preserving utility.
They proposed a variety of privacy mechanisms, including variants of Laplace, plus the novel Purkayastha and von Mises-Fisher (VMF) mechanisms. In the present work, we adopt the VMF to apply directional noise to gradients instead of (isotropic) Gaussian noise, drawing on a similar intuition that preserving the gradient directions should provide better utility.

\vspace{-0.05cm}

\paragraph{Membership Inference Attacks (MIA)} \label{sec:mia}


\citet{Shokri2016MembershipIA} showed that it is possible to infer if a sample was used to train a model by analysing the class probabilities it outputs during inference. MIA works in two steps. In the first, shadow models are trained to mimic the target model to be attacked. In the second, for each class, separate (binary) attack models are trained on the shadow models' prediction vectors to predict whether a sample was used to train the target or not (\textit{in}/\textit{out}).
The intuition is that the target model is more confident when classifying samples it has already seen, yielding different distributions between \textit{in} and \textit{out} samples.

A number of works extended this attack \citep{DBLP:conf/csfw/YeomGFJ18, DBLP:conf/ndss/Salem0HBF019}. Recently, \citet{10.1145/3548606.3560675} presented an enhanced MIA that they characterise as `population-based', using reference models to achieve a significantly higher power (true positive rate) for any (false positive rate) error, at a lower computation cost.
\citet{DBLP:conf/uss/Jayaraman019} proposed MIA as an empirical method to help in comparing privacy budgets across DP variants. We similarly use it for our comparisons.

\vspace{-0.05cm}

\paragraph{Gradient-based Reconstruction Attacks} \label{sec:gradattack}

    This attack lies within a distributed learning framework, which aims to train a neural network without centralising data. It consists of multiple clients, and each one holds its own private training set and exchange the gradients. However, it is still possible to reconstruct the training data from the gradients received.
    
    \citet{zhu-etal:2019:NeurIPS} discovered that it is possible to recover the private data in attacking neural network architectures which are twice differentiable; their attack has subsequently been referred to as the Deep Leakage from Gradients (DLG) attack. The attacker creates dummy inputs and labels, but instead of optimising the model weights, it optimises the dummy input and labels to minimise the Euclidean distance between their gradients and the gradients received from another client. Matching the gradients transforms the fake input to be similar to the real one.
    
    This attack was refined in further works. The Inverting Gradients method (\IGA), from \citet{geiping-etal:2020:NeurIPS}, maximises the cosine similarity between gradients. Thus it relies on an angle-based cost function, which should be more robust than a magnitude-based one against a trained neural network (which produces gradients with smaller magnitudes). 
    We use this widely used attack for further empirical privacy calibration.
    

\vspace{-0.1cm}

\section{The Privacy Model}
\label{sec:model}

\vspace{-0.05cm}

\begin{definition}\label{d1614-a}
(Metric differential privacy) \citep{chatzikokolakis-etal:2013:PETS} Let $\varepsilon{>}0$. A mechanism ${\mathcal M}$ on an (input) metric space $(S, d)$, where $S$ is a set and $d$ is a metric, and producing outputs over $\mathcal{Z}$, satisfies $\varepsilon d$-privacy, if for all $s, s'\in S$ and $Z \subseteq \mathcal{Z}$, 
\[
Pr ({\mathcal M}(s))[Z] \leq e^{\varepsilon d(s, s')}\times Pr({\mathcal M}(s'))[Z]~,
\]
where $Pr ({\mathcal M}(s))[Z]$ means the probability that the output of applying mechanism $\mathcal{M}$ to $s$ lies in $Z$.
\end{definition}


\Def{d1614-a} says that when two inputs $s,s'$ differ by the amount $d(s,s')$, the mechanism can make them indistinguishable up to a ratio proportional to $e^{\varepsilon d(s, s')}$. This means that when points are farther apart they become easier to distinguish. 

Metric DP, like pure $\epsilon$-DP, has straightforward composition properties: the epsilons ``add up'' under sequential composition, and for the same metric, privacy does not diminish under parallel composition; and it satisfies the data processing inequality~\citep{chatzikokolakis-etal:2013:PETS}.

\vspace{-0.05cm}

\subsection{Directional Privacy and \DirDP}

\begin{figure}[!th]{\textwidth 0mm}
		\centering
        \scriptsize
		\includegraphics[width=0.25\textwidth]{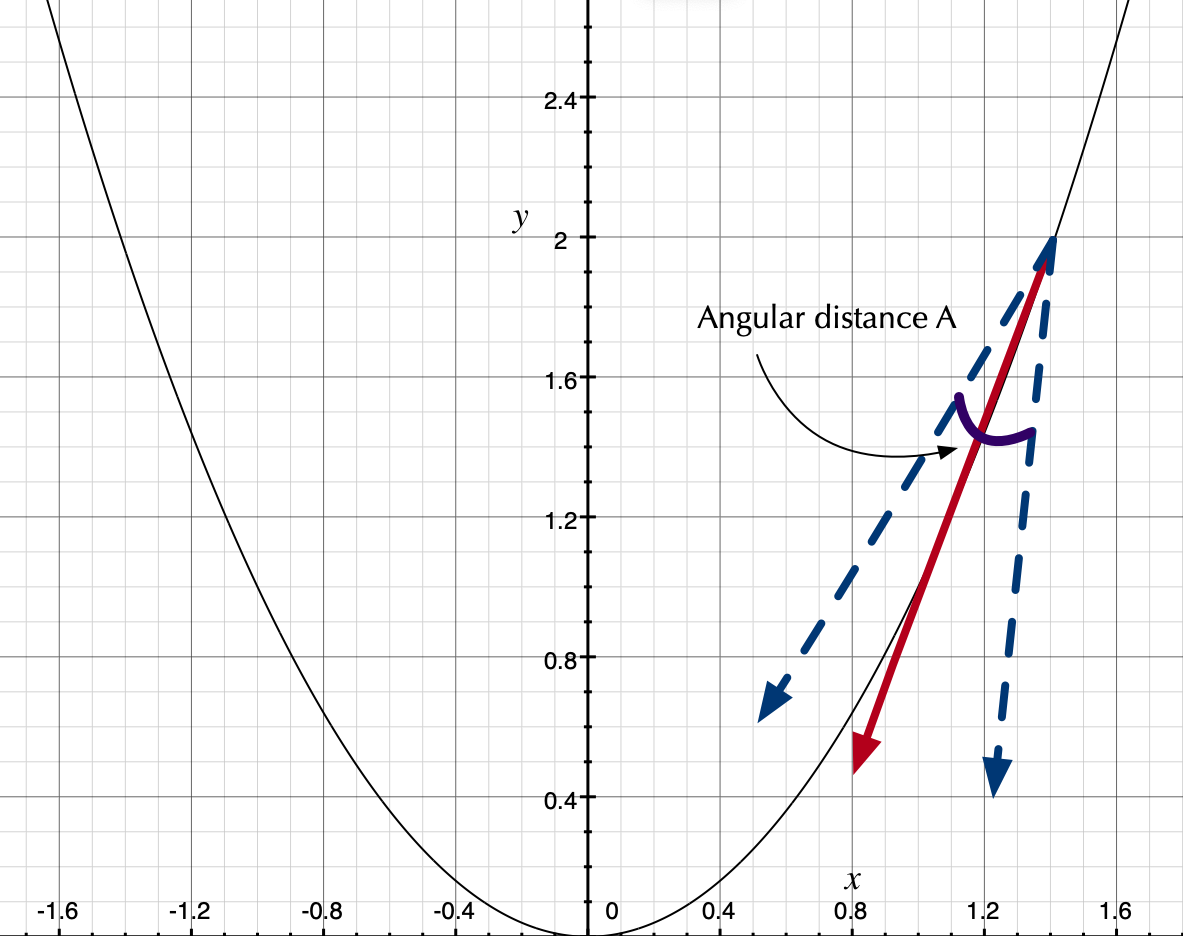}
		\caption{The red line is the unperturbed gradient, and the dotted blue lines are perturbations of angular distance $A$.}\label{f1406-a}
	\end{figure}

Gradient descent optimises the search for parameter selection that minimises the loss. An alternative method of perturbing the gradients is to use randomisation that is based on the angle of deviation from the original gradient.
To give some intuition, Figure~\ref{f1406-a} illustrates how a gradient of a convex curve can be perturbed, leading to a perturbation of the descents.


Given two vectors $v, v'$ in $\mathbb{R}^K$, the angular distance between them is  $d_\theta(v, v')= \frac{\arccos{v^Tv'}}{\| v\| \| v'\|}$. When $v,v'$ are, for example, vectors on the unit $K$-dimensional sphere, then $d_\theta$ becomes a metric. Following  \citet{weggenmann-kerschbaum:2021:CCS}, we can use this to define \emph{directional privacy}.

\begin{definition}\label{d1614}
(Directional Privacy) \citep{weggenmann-kerschbaum:2021:CCS} Let $\epsilon{>}0$. A mechanism ${\mathcal M}$ on $\mathbb{R}^K$ satisfies $\varepsilon d_\theta$-privacy, if for all $v, v'$ and $Z \subseteq \textrm{supp}\mathcal{M}$, 
\[
Pr({\mathcal M}(v))[Z] \leq e^{\varepsilon d_\theta(v, v')}\times Pr({\mathcal M}(v'))[Z]~.
\]
\end{definition}


Definition~\ref{d1614} says that when the mechanism ${\mathcal M}$ perturbs the vectors $v, v'$, the probabilities that the perturbed vectors lie within a (measurable) set $Z$ differ by a factor of $e^{\varepsilon d_\theta(v, v')}$. This means that the smaller the angular distance between the vectors $v,v'$ the more indistinguishable they will be.

\citet{weggenmann-kerschbaum:2021:CCS} introduced the von Mises-Fisher (VMF) mechanism which perturbs a vector $x$ on the $K$-dimensional unit sphere.
\begin{definition}\label{d1526}
The VMF mechanism on the $K$-dimensional unit sphere is given by the density function:
\[
\VM(\varepsilon, x)(y) ~ = ~ C_K(\varepsilon) e^{\varepsilon x^T y}~,
\]
where $\varepsilon >0$ and $C_K(\varepsilon)$ is the normalisation factor.
\end{definition}

The authors proved that the VMF mechanism satisfies $\varepsilon d_\theta$-privacy.
They also provide an analysis of the expected error of the VMF as well as sampling methods which we use later in our experiments. Importantly, the following also holds. 

\begin{theorem}[\cite{weggenmann-kerschbaum:2021:CCS}]\label{thm:d2_priv}
Let $\epsilon > 0$ and denote by $\mathbb{S}^{K-1}$ the unit sphere in $K$ dimensions. Then the VMF mechanism on $\mathbb{S}^{K-1}$ satisfies $\epsilon d_2$-privacy where $d_2$ is the Euclidean metric. That is,
\[
	\VM(\epsilon, x)(Y) ~\leq e^{\epsilon d_2(x, x')} \VM(\epsilon, x')(Y)~,
\]
for all $x, x' \in \mathbb{S}^{K-1}$ and all (measurable) $Y \subseteq \mathbb{S}^{K-1}$.
\end{theorem} 


In our application, we apply the basic mechanism $\VM$ to more complex data representations, namely where a point is a represented as convex sum of $m$ orthogonal vectors in $n$-dimensions. The standard method for doing this is to apply $m$-independent applications of the mechanism (in this case $\VM$); the differential privacy guarantee is then parametrised by $m$ as follows.

\begin{corollary}[\cite{dwork-roth:2014}]\label{c1326}
    Let $\VM$ be the mechanism defined in Definition~\ref{d1526}. Let $v, v'$ be two vectors on the unit sphere, where $v= \lambda_1u_1 + \dots \lambda_k u_m$ and  $v'= \lambda'_1u'_1 + \dots \lambda'_k u'_m$, where $u_i, u'_i$ for $1\leq i \leq m$ are vectors on the unit sphere, and $|\lambda_i|, |\lambda'_i| \leq 1$. Define $\VM^*$ to be the mechanism that applies $\VM$ independently to each of the $u_i/u'_i$ to produce  random vectors distributed respectively as: $\VM^*(\epsilon, v), \VM^*(\epsilon, v')$. Then 

    \[
     \VM^*(\epsilon, v)(Y) ~\leq e^{2\epsilon \sqrt{m}} \VM(\epsilon, v')(Y)~.
    \]
\end{corollary}

 \begin{proof}
The standard properties of differential privacy \citep{dwork-roth:2014}  result in the following relationship:        
    \[
     \VM^*(\epsilon, v)(Y) ~\leq e^{\epsilon \sum_{1\leq i\leq i'}d_2(\lambda u_i, \lambda'u_i')} \VM(\epsilon, v')(Y)~.
    \]
Observe that for any orthonormal set of vectors $u_i$ on the unit sphere, we have that  $\sum_{0\leq i \leq m}d_2(0, \lambda_iu_i)\leq \sum_{0\leq i \leq m} |\lambda_i| \sqrt m$. The result now follows using the triangle inequality of $d_2$, and that $d_2(\lambda_i u_i, \lambda'_iu_i')\leq 2$.     
    \end{proof}


There are interesting scenarios based on Corollary~\ref{c1326} which we explore in our adaptation of DP-SGD. The first is that $\VM$ is applied (once) to an $n$-dimensional vector to produce a random $n$-dimensional vector. For us, in Corollary~\ref{c1326} we would use $m=1$ to obtain an overall $2\epsilon$ for our privacy parameter. An alternative extreme is to apply noise independently to each of the components (in the way that the original DP-SGD  does); Corollary~\ref{c1326} then gives a $\epsilon \sqrt n$ privacy budget. An interesting hybrid scenario, not available for the Gaussian distribution but available for the $\VM$ mechanism, is to partition the $n$-dimensional components into $m$-orthogonal components and to apply $\VM$ independently to each of those components; in this case, we obtain the $\epsilon\sqrt m$ privacy budget as in Corollary~\ref{c1326}. As explained in our experiments, we found that for some of the datasets, this was a useful generalisation for the purposes of efficiency.


Using \Def{d1526}  we can now design a new DP-SGD algorithm using the VMF mechanism which perturbs the \emph{directions} of the gradients computed in SGD. This algorithm, which we call \DirDP, is depicted in \Alg{alg:sgd2}. It is modified from the original DP-SGD (blue text) in three ways. First, we fix $C$ to $1$. Then (line 7), instead of clipping the gradients, we scale the gradients to the clip length (i.e. 1). Finally (line 9), instead of adding a noisy vector to the gradient, the VMF mechanism generates a new noisy vector directly based on its input.

\begin{algorithm}[!th]
\small
\caption{\DirDP with von Mises-Fisher noise}\label{alg:sgd2}
\begin{algorithmic}[1]
\State \textbf{Input:} Samples $\{x_1,\ldots,x_N\}$, loss function $\mathcal{L}(\theta) = \frac{1}{N} \sum_i \mathcal{L}(\theta, x_i)$. Parameters: learning rate $\eta_t$, noise scale $\sigma$, group size $L$, {\color{blue}gradient norm bound $C = 1$.}
\State \textbf{Initialise} $\theta_0$ randomly
\For{$t \in T$}
   \Comment{Take a random batch}
   \State $L_t \gets $ random sample of $L$ indices from $1{\ldots}N$
   \For{$i \in L_t$}
         \Comment{Compute gradient vector}
   	\State $\mathbf{g}_t(x_i) \gets  \nabla_{\theta_t} \mathcal{L}(\theta_t, x_i)$
	\Comment{Scale gradient vector}
	{\color{blue}\State $\gbar_t(x_i) \gets \nicefrac{\mathbf{g}_t(x_i)}{\frac{\| \mathbf{g}_t(x_i)\|_2}{C} }$}
  \EndFor  
   \Comment{Add noise}
   \State {\color{blue}  $ \gtilde_t \gets \frac{1}{L} \sum_{i} \VM(\sigma, \gbar_t(x_i)) $}
   \Comment{Descent}
   \State $\theta_{t+1} \gets \theta_t - \eta_t \gtilde_t$
\EndFor
\State \textbf{Output} $\theta_T$ 
\end{algorithmic}
\end{algorithm}




\Alg{alg:sgd2} satisfies $\epsilon$-DP and $\epsilon d_\theta$-privacy in terms of indistinguishability of batches used in the learning, viz that if two batches (composed of data points $x_i$) differ in only one point, then they are probabilistically indistinguishable.

\begin{theorem}\label{l1647}
Denote by $B = [v_1, \dots v_n]$ , and $B' = [v'_1, \dots v'_n]$ two batches of vectors (gradients). If batch $B'$ differs from $B$ in at most one component vector, then \Alg{alg:sgd2} satisfies $\sigma d_2$-privacy wrt.\ batches, namely that:

\begin{equation}\label{e1720}
Pr({\mathcal VM}(B) \in Z) \leq Pr({\mathcal VM}(B') \in Z) \times e^{\sigma d_2(B, B')} ~,
\end{equation}

 $Z$ is a (measurable) set of vectors, $Pr({\mathcal VM}(B) \in Z)$ is the probability that the output vector lies in $Z$ and (abusing notation) $d_2(B, B') = \max_{B \sim B'} d_2(v, v')$ is the maximum Euclidean distance between all pairs $v\in B, v'\in B'$.

\end{theorem}

\begin{proof}
Line 7 of \Alg{alg:sgd2} ensures each vector in $B, B'$ lies on the unit sphere (since $C = 1$) and line 9 applies VMF parametrised by $\sigma$.
Applying \Thm{thm:d2_priv} to every vector in $B, B'$ yields \Eqn{e1720} since the (parallel) composition of $d_2$-privacy mechanisms gives a guarantee
with $Pr({\mathcal VM}(B) \in Z) \leq Pr({\mathcal VM}(B') \in Z) \times e^{\sigma \sum_{1\leq i\leq n}d_2(v_i, v_i')}$; this reduces to \Eqn{e1720} since all but one of the distances $d_2(v_i, v'_i)=0$, by assumption.
 The averaging in line 9 is a  post-processing step which, by the data processing inequality property of $d$-privacy does not decrease the privacy guarantee \citep{fernandes:22:CSF}.
 \end{proof}

Observe that \Alg{alg:sgd2} assumes we apply the $\VM$ mechanism to the whole gradient; as mentioned above, in our experiments we sometimes partition the $n$-dimensional space. We proceed though to prove a privacy guarantee assuming  \Alg{alg:sgd2} applies $\VM$ without partitioning. 

\begin{corollary}\label{cor:epsdp}
\Alg{alg:sgd2} satisfies $\epsilon$-DP wrt\ adjacent training sets $D, D'$.

\end{corollary}

\begin{proof}
Observe that $\max_{B \sim B'} d_2(v, v') = 2$ since $v, v'$ lie on the unit sphere. $\epsilon$-DP on batches follows from choosing $\sigma=\frac{\epsilon}{2}$ which is the standard DP-SGD tuning from \citet{song-etal:2013}. Since batches are disjoint, the result follows by parallel composition for adjacent training sets $D, D'$. 
\end{proof}

\begin{corollary}\label{cor:edpriv}
\Alg{alg:sgd2} satisfies $\epsilon d_\theta$-privacy.

\end{corollary}

\begin{proof}
Follows from the fact that $d_2$-privacy implies $d_\theta$-privacy (since $d_2 \leq d_\theta$ pointwise on the unit sphere), using $d_\theta$ reasoning in \Thm{l1647} and using the same $\sigma$ tuning as per Cor~\ref{cor:epsdp}.
\end{proof}

Note that the epsilons in Cor~\ref{cor:epsdp} and Cor~\ref{cor:edpriv} are not comparable as they represent different notions of privacy.

We remark that by \emph{scaling} rather than clipping the gradients, we also protect against privacy leaks caused when the gradient length is less than $C$.  Information may or may not be leaked by knowing the length of the gradient, but we remove this possibility by scaling rather than clipping.




\section{Experimental Setup}
\label{sec:exper}

We compare the accuracy of different neural networks when they are trained with DP against baselines without privacy guarantees, as by \citet{abadi-etal:2016:CCS, li-etal:2022:ICLR}. We take an empirical approach to calibrating the respective $\epsilon_G$ and $\epsilon_V$: we measure how \DirDP prevents MIA-R and gradient-based reconstruction attacks at equivalent levels of utility.

We use the training and test sets from \textbf{Fashion-MNIST} \citep{Xiao2017FashionMNISTAN} and \textbf{CIFAR} \citep{Krizhevsky09learningmultiple} datasets. We also use the Labeled Faces in the Wild (\textbf{LFW}) dataset \citep{Huang07labeledfaces}, but given its large number of classes, we follow \citet{wei-etal:2020:ESORICS} to keep only classes with at least 14 objects. This reduced the number of classes to 106 and samples to 3,737. Classes are still imbalanced, therefore we under-sampled the majority classes by randomly picking objects so that all classes have 14 samples. This reduced the dataset further to 14*106 = 1,484 instances. Finally, we split it into training (80\%, or 1,113 samples) and test (20\%, or 371 samples) sets. 

Given its many classes and few instances, one might expect low accuracies for LFW, which the added noise might reduce to near-zero levels, obscuring differences among different types and levels of noise. Thus, we report the Top-10 accuracy, where a prediction is correct if the true label is amongst any of the top 10 predictions with the highest score.

We broadly follow the setup of \citet{scheliga-etal:2022:WACV}. We use \textbf{LeNet}, the original convolutional neural network (CNN) proposed by \citet{bb72cefb6cc34854965b753d1ce10cbd}, and a simple Multilayer Perceptron (\textbf{MLP}) with 2 layers, which are feedforward neural networks \citep{GoodBengCour16}. 
 \citet{scheliga-etal:2022:WACV} include MLPs as they note that \citet{geiping-etal:2020:NeurIPS} provide a theoretical proof that in fully connected networks, their \IGA attack can uniquely reconstruct the input from the gradients.

We selected values going from `small' ($\epsilon_G \leq 1$) to the common largest value of 8 that a number of works \citep[for example]{abadi-etal:2016:CCS,de-etal:2022} have selected. We also added  $\epsilon_G = 50$ as a relatively small amount of noise, although this is larger than in many other works, for calibration purposes.


    There is no prior work with VMF as a guide for selecting $\epsilon_V$. Based on preliminary experiments, we found a range of changes to utility in $\epsilon_V \in \{ 1, 5, 10, 50, 500 \}$; we also included $\epsilon_V = 300,000$, which hardly shifts gradients, to investigate the effects of negligible noise. Hyperparameters are in Table \ref{tab:app_hyperparameters}.

    \begin{table}[h]\label{tab:hyperparameters}
            \centering
            \tiny
            \begin{tabular}{ccccccc}
            \hline
            \textbf{Model} & 
            \textbf{Dataset} &
            \textbf{Batch size} &
            \textbf{LR} &
            \textbf{Epochs} &
            \textbf{Optim.} &
            \textbf{Momentum} \\ \hline
             \multirow{3}{*}{Lenet} & CIFAR & 512 & 0.001 & 90 & Adam  &0.0\\
              & FashionMNIST & 100 & 0.001 & 30 & Adam &0.0\\
              & LFW &  128 & 0.05  & 60 &  SGD & 0.9\\ \cline{2-7}
             \multirow{3}{*}{MLP} & CIFAR & 128 & 0.001 & 25 & Adam &0.0\\
              & FashionMNIST & 256 & 0.01 &30 & Adam  &0.0 \\
              & LFW & 256  &  0.001 &  60 & Adam   &0.9\\
              \hline
                \multirow{2}{*}{Lenet} & CIFAR & 8 & 0.01 & 60 & Adam &0.0\\
              & FashionMNIST & 8 & 0.001 & 10 & Adam &0.0\\ \cline{2-7}
             \multirow{2}{*}{MLP} & CIFAR & 256 & 0.001 & 25 & Adam  &0.0\\
              & FashionMNIST & 8 & 0.1 &30 & Adam  &0.0 \\
             \hline
            \end{tabular}
            \caption{\textmd{Hyperparameters set. Bottom ones were used for MIA-\textbf{R}.}}
        \label{tab:app_hyperparameters}
    \end{table}



\section{Results}
\label{sec:results}

\subsection{Utility experiments}

    We compare the models after they are trained with and without DP guarantees in classification tasks across different datasets. Table \ref{tab:utility_acc} shows the test accuracy after training on their respective training sets.

    \begin{table}[t]
        \centering
        \scriptsize
            \begin{tabular}{lccccc}\hline
            & &  FMNIST & CIFAR 10 &CIFAR 100 & LFW   \\ \hline
            \textbf{Model}  & \textbf{$\epsilon_G$, $\epsilon_V$} & \multicolumn{3}{c}{\textbf{Accuracy}} & \textbf{Top-10}  \\
            \hline
            LeNet & -& 87.7  & 52.2 & 24.4 & 32.0  \\
            MLP  & -&   84.7 & 45.8 &  16.5 & 15.6  \\ \hline
            \multirow{6}{*}{\makecell{LeNet \\ Gauss}} & 0.8 &  81.2  & 36.7 &  4.9 &   8.9  \\
             &   1.0 &  81.7  & 37.4 & 5.5  & 9.6   \\
             &   2.0 &  82.3  & 41.5 &  7.8 &  8.7  \\
             &   3.0 &  83.1  & 43.2 & 9.0  &  8.7  \\
             &   8.0 &  83.9  & 46.6 &  12.0 & 13.2   \\
              &   50.0 &  85.1  & 48.9 & 15.3  &  15.2  \\
             \hline
             
             \multirow{6}{*}{\makecell{LeNet \\ VMF}} & 1 &  81.5   &  50.7   & 21.4 & 11.8   \\
             &   5 &  81.3   & 51.3  & 20.7  &  11.2  \\
             &   10 & 81.8  & 50.5  &  21.0 &  11.2  \\
             &   50 & 81.9  & 51.4  & 21.2 &   13.0 \\
             &   500 & 82.9 &  51.4  & 22.2 &  13.9  \\
             &   300k &  83.9 & 51.6  & 25.8  &  14.3  \\\hline
             
             \multirow{6}{*}{\makecell{MLP \\ Gauss}} & 0.8 &   79.6 & 32.0 & 5.6  &  10.5  \\
             &   1.0 & 79.8   & 32.7 & 5.9  &   10.5 \\
             &   2.0 &  81.2  & 34.6 & 7.0  &  10.3  \\
             &   3.0 & 81.7   & 35.3 & 7.7  & 9.4   \\
             &   8.0 &  82.9  & 36.9 &  8.9 &  9.6  \\
              &   50.0 &  84.1  & 39.1 & 10.6  &  10.7  \\
             \hline
             
             \multirow{6}{*}{\makecell{MLP \\ VMF}} & 1 &   84.9  & 42.1   & 13.4 & 14.3   \\
             &   5 &  84.2   & 41.9  & 13.6  & 15.2   \\
             &   10 & 84.4  & 42.3  & 13.8  & 14.7   \\
             &  50 &  84.5  &  42.9  &  14.6&  15.2  \\
             &   500 &  85.1  & 43.3  &  15.8 &  18.3  \\
             &   300k &  85.3  & 44.8 &  17.0 &  27.1  \\\hline
             
            \end{tabular}%
            \caption{\textmd{Classification accuracy under different DP settings for the test sets. } }\label{tab:utility_acc}
    \end{table}

    The first two rows show the test set accuracy of baseline models without any DP mechanism. The remaining rows bring results of the same models, but with different DP mechanisms (Gauss and VMF) and with different values of their respective privacy parameters $\epsilon_G$ and $\epsilon_V$.

    With few exceptions, more noise reduces utility: considering $\epsilon_G$ and $\epsilon_V$ as privacy budgets, the higher they are, the less privacy should be retained, thus increasing the accuracy.
    
    Overall, in terms of non-private models, relative performance on the datasets and tasks is as expected: Fashion-MNIST is the easiest one. Even small values of $\epsilon$, like $\epsilon_G = 0.8$ don't seem to drastically affect the utility of the model compared to the non-private baselines. However, we observe other trends when looking at the other datasets. 

    For the CIFAR dataset, VMF noise leads to much smaller reductions in utility for Lenet and MLP; even the largest $\epsilon_G = 50$ does not reach the accuracy of the smallest VMF-$\epsilon_V$. For LFW, there is little difference in performance across the models: the accuracy drop is sharper for Lenet, whereas the bigger $\epsilon$ approach is closer to the original accuracy.

    We also observe some cases when the neural network is trained with VMF noise, its accuracy is even marginally higher than the baseline without any DP guarantees for a large or very large value of $\epsilon$ (e.g. MLP on Fashion-MNIST, FLW and CIFAR100). We hypothesise that the noise also acts as a regularisation factor that prevents overfitting, which can explain the modest performance gain on the test set. In fact, the use of noise as a regularisation technique has been studied by \citet{li-liu:2020, li-liu:2021}, but with Gaussian noise.

\begin{figure*}[ht!]
\centering
\small
\begin{tabular}{ccccc}

    \subfloat[Lenet on Cifar]{\includegraphics[width=0.19\textwidth]{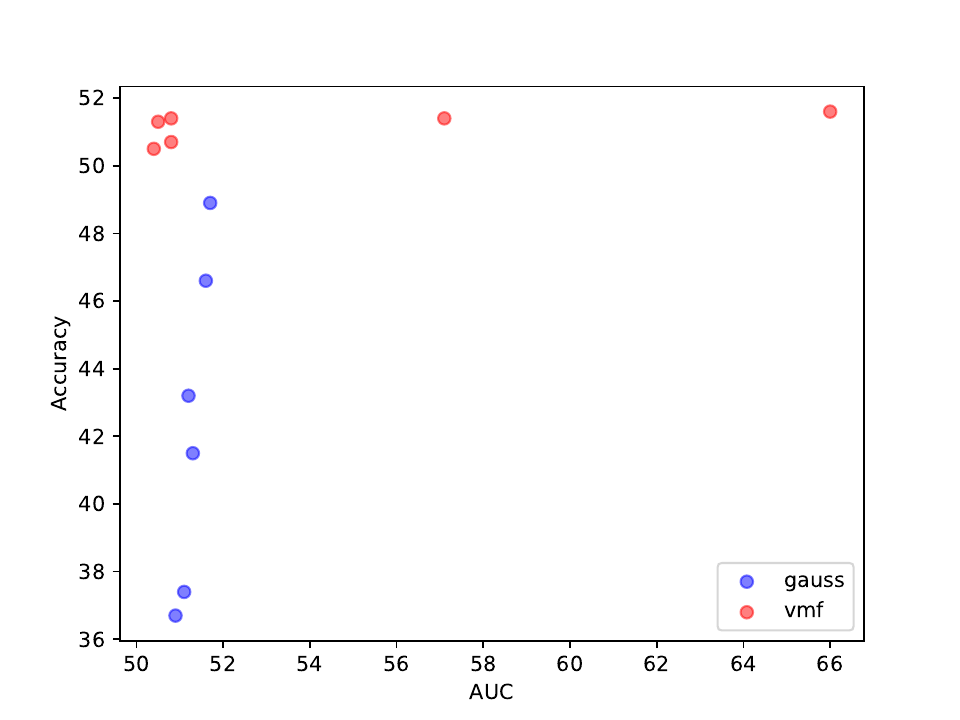}} &
    
    \subfloat[MLP on Cifar]{\includegraphics[width=0.19\textwidth]{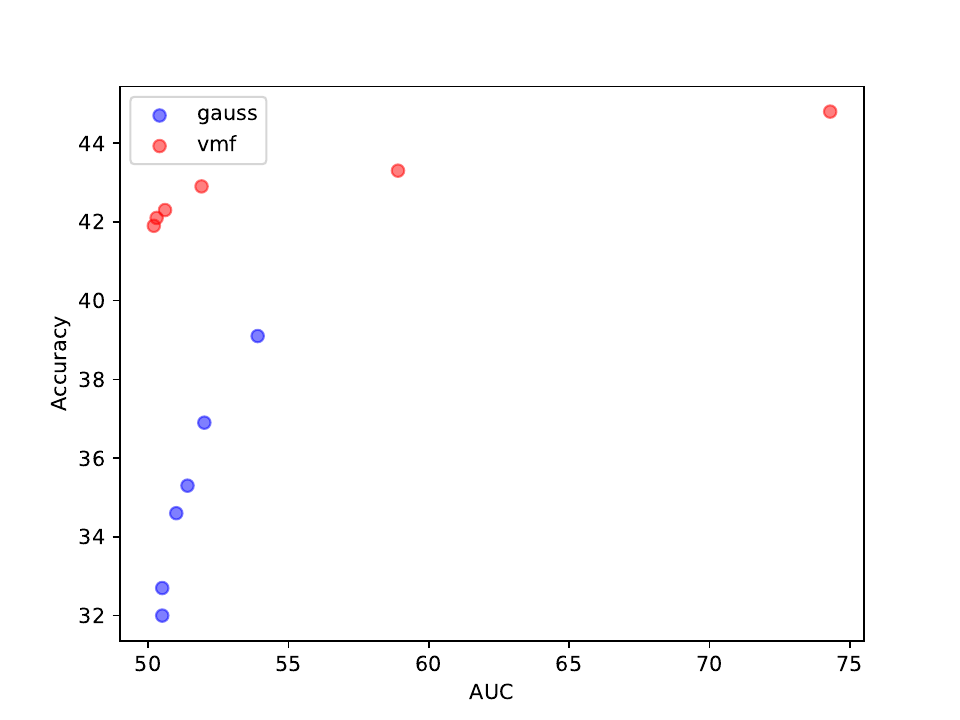}} &
    
    \subfloat[Lenet on FMNIST]{\includegraphics[width=0.19\textwidth]{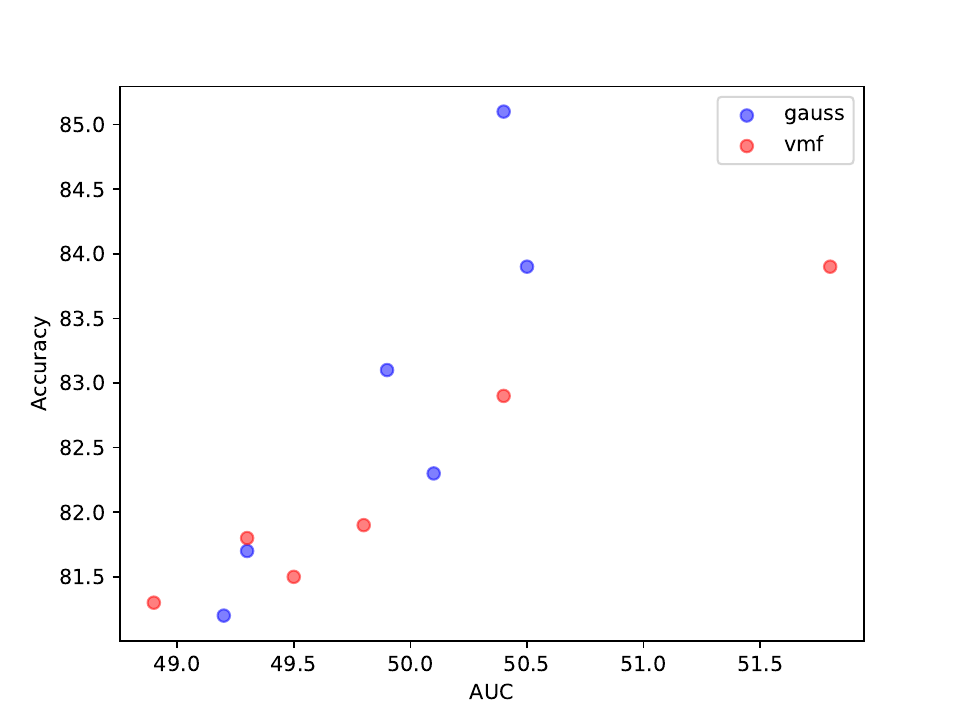}} &
    
    \subfloat[MLP on FMNIST]{\includegraphics[width=0.19\textwidth]{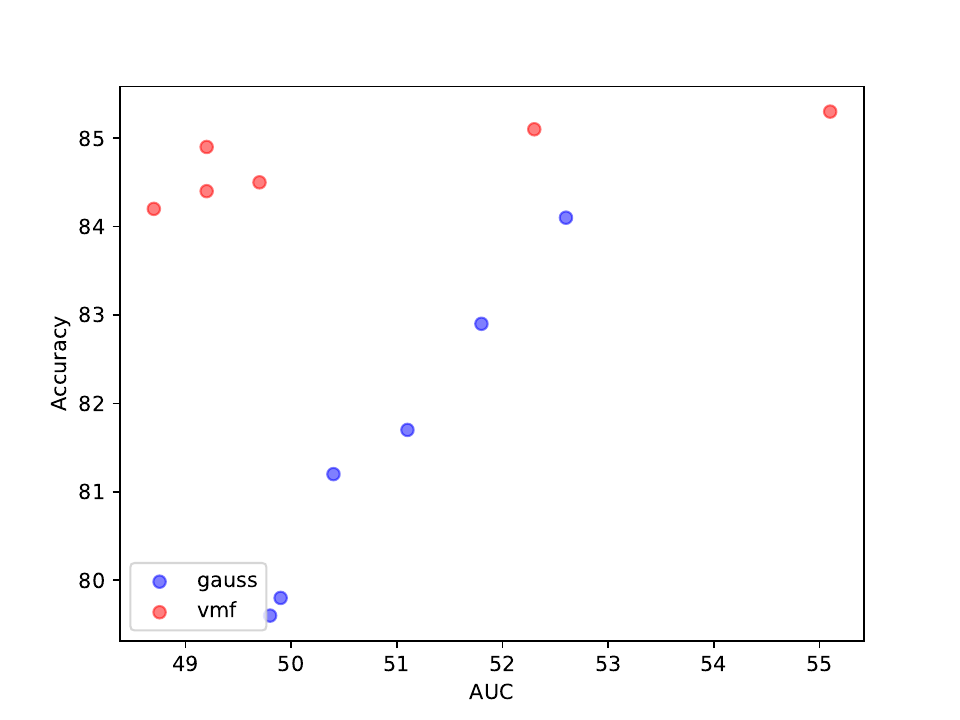}} \\
    
    \subfloat[Lenet on Cifar]{\includegraphics[width=0.19\textwidth]{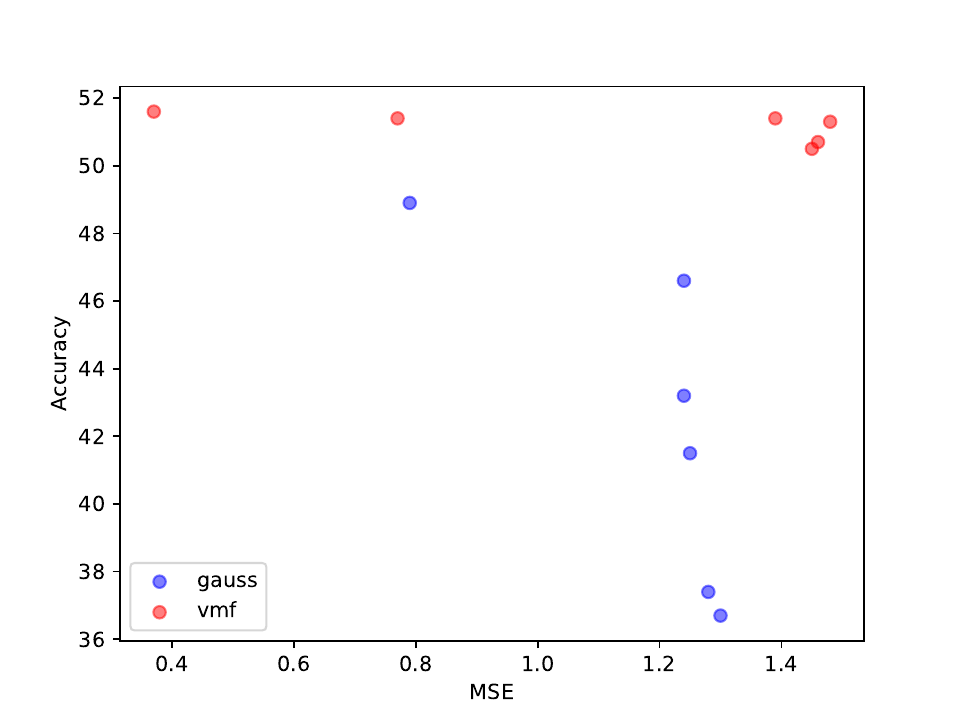}}  &

    \subfloat[MLP on Cifar]{\includegraphics[width=0.19\textwidth]{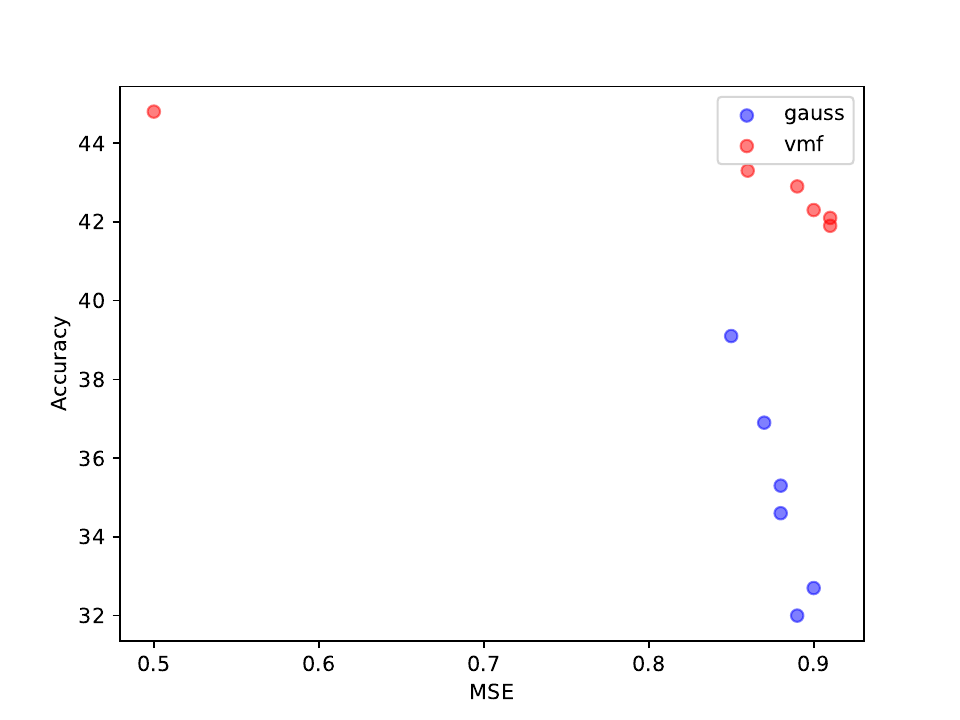}} &
     
    \subfloat[Lenet on FMNIST]{\includegraphics[width=0.19\textwidth]{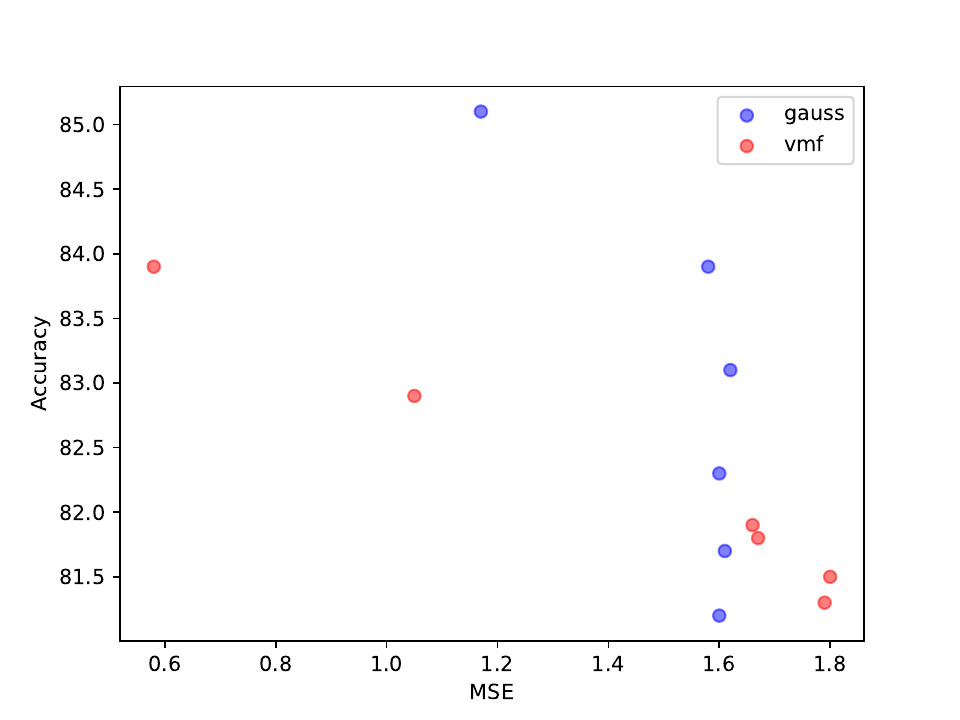}} &
    
    \subfloat[MLP on FMNIST]{\includegraphics[width=0.19\textwidth]{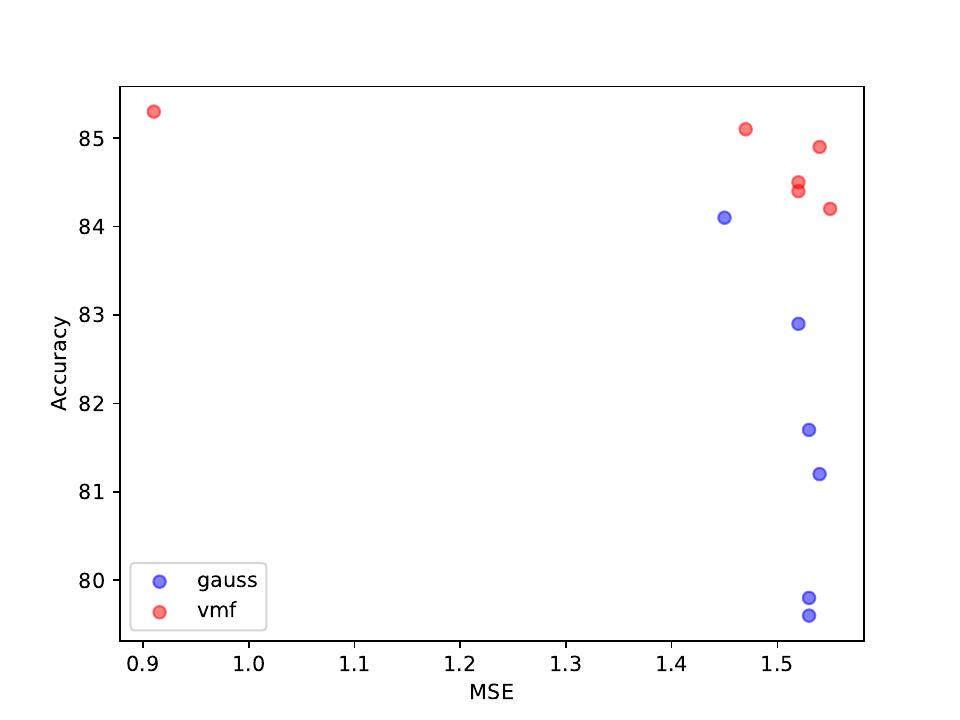}} \\ 

\end{tabular}
\caption{\textmd{Tradeoffs between utility (accuracy) and privacy (AUC for MIA or MSE for reconstruction attacks).}}\label{fig:scatter_tradeoff}
\end{figure*}

\vspace{-0.05cm}

\subsection{Enhanced MIA}

\begin{table}[t]
    \centering
    \scriptsize
        \begin{tabular}{lcccc}\hline
        \textbf{Model} & \textbf{$\epsilon_G$, $\epsilon_V$} & \textbf{F-MNIST} &\textbf{CIFAR10}  \\
        \hline
        LeNet & -- & 54.8    & 78.5 \\
        MLP & -- &   61.1  & 82.0  \\ \hline
        \multirow{6}{*}{\makecell{LeNet \\ Gauss}} 
           & 0.8 & 49.2 & 50.9  \\
        &   1.0 & 49.3 &  51.1  \\
        &   2.0 &  50.1  &  51.3  \\
        &   3.0 &  49.9  & 51.2 \\
        &   8.0 & 50.5 & 51.6  \\
        &   50.0 & 50.4 & 51.7  \\
        \hline
         
         \multirow{6}{*}{\makecell{LeNet \\ VMF}} 
            & 1.0 & 49.5 &  50.8  \\
        &   5.0 & 48.9 & 50.5 \\
        &   10.0 & 49.3 & 50.4 \\
        &   50.0 & 49.8 &  50.8 \\
        &   500.0 & 50.4 & 57.1 \\
        &   300k & 51.8 &  66.0 \\
        \hline
        
         \multirow{6}{*}{\makecell{MLP \\ Gauss}} 
           & 0.8 & 49.8 &  50.5 \\
          &   1.0 &  49.9  & 50.5  \\
          &   2.0 & 50.4   &   51.0  \\
          &   3.0 & 51.1   & 51.4    \\
          &   8.0 &  51.8  &    52.0  \\
          &   50.0 &   52.6   &  53.9 \\
         \hline
         
         \multirow{6}{*}{\makecell{MLP \\ VMF}} 
            & 1.0 & 49.2 & 50.3  \\
         &   5.0 & 48.7 &   50.2 \\
         &   10.0 & 49.2 &  50.6  \\
          &  50.0 &   49.7  &  51.9  \\
        &   500.0 & 52.3 & 58.9 \\
          &   300k & 55.1   &  74.3  \\
         \hline
         
        \end{tabular}%
        \caption{\textmd{AUC attack scores under different DP settings for the MIA-R, comparing Gaussian and VMF noise. Lower scores mean better privacy.} }\label{tab:miar_attack_auc}
    \end{table}

Table ~\ref{tab:miar_attack_auc} gives the results for the enhanced MIA-\textbf{R} attack, for both Gaussian and VMF noise.  We left the LFW dataset out of this set of experiments due to its limited size to be split into several training partitions for the reference models. 

The attack was particularly successful against CIFAR, with a large improvement over chance against the non-private models.  Comparing Gaussian against VMF on LeNet, we see that even the largest $\epsilon_V$ reduces the attack success to less than the $\epsilon_G = 1$.

For Fashion-MNIST, the baseline models without any privacy guarantee leak less information than those trained with CIFAR. Thus, unsurprisingly, most noise settings make the attacks similar to random chance, except for some cases where $\epsilon$ is very large, particularly with MLP. MIA-\textbf{R} thus cannot help us calibrate $\epsilon$ here.

\vspace{-0.05cm}

\subsection{Defence against Gradient Leakage Attacks}

    \begin{table}[t]
        \centering
        \scriptsize
            \begin{tabular}{ccccc}\hline
            \textbf{Model} & \textbf{$\epsilon_G / \epsilon_V$} & \textbf{F-MNIST} & \textbf{CIFAR} &\textbf{LFW}   \\
            \hline
             LeNet & -- &  0.65    & 0.36  & 0.68 \\
             MLP & -- &   0.37 &  0.18 & 0.31  \\
             \hline
            \multirow{6}{*}{\makecell{LeNet \\ Gauss}}  & 0.8&   1.60   &   1.30   & 1.51 \\
             & 1.0  &   1.61  & 1.28 & 1.51 \\
             & 2.0  &   1.60   & 1.25  & 0.49 \\
             & 3.0 &    1.62   &  1.24 & 1.51 \\
             & 8.0  &   1.58  &  1.24 & 1.39 \\
              &  50  &   1.17  &  0.79 & 1.09 \\
             \hline
             
             \multirow{6}{*}{\makecell{LeNet \\ VMF}}  & 1  &   1.80  & 1.46  & 1.51 \\
             & 5&  1.79  &1.48  & 1.56 \\
             & 10 &  1.67   &  1.45 & 1.53 \\
             & 50 &   1.66  & 1.39  & 1.29 \\
             & 500 &  1.05   & 0.77  & 1.07 \\
             & 300k &    0.58 &  0.37 & 0.75 \\\hline

             \multirow{6}{*}{\makecell{MLP \\ Gauss}} & 0.8&  1.53  & 0.89 & 1.37 \\
             & 1.0& 1.53  &  0.90 &  1.38\\
             & 2.0 & 1.54  & 0.88  & 1.37  \\
             & 3.0  &   1.53  &  0.88  &1.37 \\
             & 8.0 & 1.52  & 0.87  & 1.36 \\
              &  50 &   1.45  & 0.85  & 1.32 \\
             \hline
             
             \multirow{6}{*}{\makecell{MLP \\ VMF}}  & 1 &   1.54  &   0.91   & 1.38 \\
             &  5&   1.55  & 0.91  & 1.39 \\
             &  10 &   1.52  & 0.90  & 1.36 \\
             &  50 &  1.52    & 0.89    & 1.38 \\
             &  500 &  1.47   & 0.86  & 1.33 \\
             &  300k &   0.91  &  0.50    &0.86  \\\hline
             
            \end{tabular}%
            \caption{IGA \textmd{reconstruction MSE} for LeNet and MLP.}\label{tab:attack_iga}
    \end{table}

Table \ref{tab:attack_iga} shows the results of the IGA (Inversion Gradient Attack) reconstruction attack against LeNet and MLP models, measuring the ability of an attacker to reconstruct the input data from the model's output probability distribution.

The MSE metric can be seen as the amount of noise that was kept in the resulting image after the attack was completed. As it is unbounded, we report its median to avoid a few large MSE values from dominating the average. 

For the most part, the reconstructions under Gaussian and VMF are noisier than the non-private ones.  The exception is the large $\epsilon_V = 300k$  we use for understanding the range of $\epsilon_V$s, where the MSE scores are only a little larger than for the non-private models. This indicates that both LeNet and MLP models can be vulnerable to the IGA attack if the noise is too small.

We also see that when Lenet's gradients receive Gaussian noise, the error drops sharply as $\epsilon$ increases, mainly for Fashion-MNIST and CIFAR images. For the remaining experiments, the range of values is much lower, even though the same trend can be observed in some cases, as in the MLP with Gaussian noise for Fashion-MNIST.

Overall, the MSE values for Gaussian and VMF are similar (e.g. for CIFAR MLP, both are around 0.9 for their ranges, with the exception of $\epsilon_V = 300k$ as noted above).  Where this is not the case, it is broadly in favour of VMF: for example, for the smaller $\epsilon$ for LeNet on Fashion MNIST and CIFAR, the noise added by VMF is greater than Gaussian.  

\subsection{Calibration Outcome}

For each dataset, both methods of calibration indicate that the nominal values of $\epsilon_G$ and $\epsilon_V$ are comparable for ranges $0.8 \leq \epsilon_G \leq 50$ and $1 \leq \epsilon_V \leq 50$, in those cases where VMF is not better.  To facilitate comparison, Table~\ref{tab:privacy_attack_comparison} shows the differences between attack success rates under MIA-R, and the MSE scores for IGA.  In the former case, negative scores are better for VMF; in the latter, positive scores are better for VMF.  It is only for the LeNet on Fashion-MNIST for $\epsilon = 1$ under MIA-R where Gaussian is better, and in this case, MIA is only at chance, yielding no difference between the mechanisms.



    \begin{table}[h!]
        \centering
        \scriptsize
            \begin{tabular}{ccccc}\hline
            & & \multicolumn{3}{c}{$\Delta_{\mbox{MIA-R}} (\downarrow)$} \\ 
            \textbf{Model} & \textbf{$\epsilon_G / \epsilon_V$} & \textbf{F-MNIST} & \textbf{CIFAR} & \\ \hline
            \multirow{2}{*}{LeNet} 
            	& 1	& 0.2 & -0.3 \\
                & 50 & -0.6 & -0.9 \\
                \hline

                \multirow{2}{*}{MLP} 
        & 1 & -0.7 & -0.2 \\
        & 50 & -2.9 & -2 \\ \hline
                
            & & \multicolumn{3}{c}{$\Delta_{\mbox{IGA}} (\uparrow)$} \\ 
            &&\textbf{F-MNIST} & \textbf{CIFAR} &\textbf{LFW}   \\
             \hline
            \multirow{2}{*}{LeNet} & 1 & 0.19 & 0.18 & 0\\
	           & 50 & 0.49 & 0.6 & 0.2\\	

        \hline

        \multirow{2}{*}{MLP} 
        & 1  & 0.01 & 0.01 & 0\\	
	& 50 & 0.07 & 0.04 & 0.06\\

        \hline             
            \end{tabular}%
        \caption{Comparison of empirical privacy. $\Delta_{\mbox{MIA-R}}$ indicates MIA-R attack success under Gaussian noise minus attack success under VMF noise; negative is better for VMF.
    $\Delta_{\mbox{IGA}}$ indicates IGA MSE under Gaussian noise minus MSE under VMF noise; positive is better for VMF.
    }\label{tab:privacy_attack_comparison}
    \end{table}

Figure \ref{fig:scatter_tradeoff} shows scatter plots with the trade-off between privacy (one for MSE and another for AUC) and utility (accuracy). 
In (a)-(d), top left is best (high utility, low attack success); in (e)-(h), top right is best (high utility, high MSE obscuring reconstruction).
\textbf{We can see that, for most cases where both mechanisms achieve comparable privacy} (blue and red dots are over similar values on x-axis), \textbf{the utility is better for VMF} (y-axis). The exception is Lenet for Fashion-MNIST for both attacks, where the trade-off is blurred with no clear winner.

\vspace{-0.2cm}

\section{Conclusions}
\label{sec:conclusions}

\vspace{-0.15cm}

We defined \DirDP for directional privacy by adding noise to the gradients. This problem is particularly relevant because several studies have shown that private training data can be discovered under certain settings, such as sharing gradients.

Our mechanism provides both $\epsilon d$-privacy and $\epsilon$-DP guarantees rather then $(\epsilon, \delta)$-DP. 
Because the $\epsilon$s are not analytically comparable across frameworks, we analyse membership inference attacks and gradient-based reconstruction for privacy leakage.  We show that the enhanced MIA of \citet{10.1145/3548606.3560675} and the Inverted Gradient Attack of \citet{geiping-etal:2020:NeurIPS} are useful for calibrating nominal values of $\epsilon$ across standard DP and metric-DP.  Our experiments showed that \DirDP can provide better utility than standard DP for similar levels of privacy. 

\DirDP is based on the VMF distribution and can be computationally expensive in high-dimensions. Future work includes optimising the mechanism in the same way that DP-SGD has seen effort. Also, experiments were restricted to image datasets, and we plan to explore domains such as NLP.

\vspace{-0.3cm}

\section*{Declaration of competing interest}
\vspace{-0.15cm}
The authors declare that they have no known competing financial interests or personal relationships that could have appeared to influence the work reported in this paper.

\vspace{-0.1cm}

\section*{Data availability}
Data will be made available on request.

\vspace{-0.1cm}

\section*{Acknowledgements}
This project was undertaken with the assistance of resources and services from the National Computational Infrastructure (NCI), supported by the Australian Government. This project also received funding from Oracle Labs and by the International Macquarie University Research Excellence Scholarship.  This paper is under consideration at Pattern Recognition Letters.

\bibliographystyle{apalike}
\bibliography{main}

\end{document}